\documentclass[10pt,twocolumn,letterpaper]{article}

\usepackage{iccv}
\usepackage{times}
\usepackage{epsfig}
\usepackage{graphicx}
\usepackage{amsmath}
\usepackage{amssymb}
\usepackage{pifont}
\usepackage{multirow}
\usepackage{capt-of}
\usepackage{todonotes}
\usepackage[pagebackref=true,breaklinks=true,letterpaper=true,colorlinks,bookmarks=false]{hyperref}

\usepackage{colortbl}
\definecolor{Gray}{gray}{0.95}
\definecolor{Orange}{rgb}{1, 0.94, 0.9}
\newlength\savedwidth
\newcommand{\whline}{\noalign{\global\savedwidth\arrayrulewidth
                            \global\arrayrulewidth 1.5pt}%
                   \hline
                   \noalign{\global\arrayrulewidth\savedwidth}}
\iccvfinalcopy

\def\iccvPaperID{2392} % *** Enter the ICCV Paper ID here

\ificcvfinal\fi
\begin{document}

%%%%%%%%% TITLE
%\title{Learning to Foveate for Efficient Transmission}
\title{Cost-Aware Fine-Grained Recognition for IoTs Based on Sequential Fixations}

\author{Hanxiao Wang, Venkatesh Saligrama, Stan Sclaroff, Vitaly Ablavsky\\
Boston University\\
{\tt \small \{hxw,srv,sclaroff,ablavsky\}@bu.edu}}

\maketitle
%\thispagest

\begin{abstract}
\vspace{-5pt}
We consider the problem of fine-grained classification on an edge camera device that has limited power. The edge device must sparingly interact with the cloud to minimize communication bits to conserve power, and the cloud upon receiving the edge inputs returns a classification label. %Our goal is to achieve state-of-art accuracy with minimal communications.
%distributively screening redundancies in device information and transmitting informative data.
%
%by Internet-of-Things (IoT) scenario where low-power ``edge devices'' perform sensing (e.g., capture images/video)
%but must transmit this data to the cloud for processing. 
%
%Current deep network architectures are ill-suited for this distributed task, and attention-based methods require the entire high-acuity image to be available up-front.
%
%As an exemplar, we consider fine-grained image recognition problem with a low-powered camera-edge device that must sparingly interact with the cloud, and the cloud upon receiving the edge inputs must return a classification label. %
To deal with fine-grained classification, we adopt the perspective of sequential fixation with a foveated field-of-view to model cloud-edge interactions. We propose a novel deep reinforcement learning-based foveation model, DRIFT, that sequentially generates and recognizes mixed-acuity images. Training of DRIFT requires only image-level category labels and encourages fixations to contain task-relevant information, while maintaining data efficiency. Specifically, we train a foveation actor network with a novel Deep Deterministic Policy Gradient by Conditioned Critic and Coaching (DDPGC3) algorithm. In addition,  we propose to shape the reward to provide informative feedback after each fixation to better guide RL training. We demonstrate the effectiveness of DRIFT on this task by evaluating on five fine-grained classification benchmark datasets, and show that the proposed approach achieves state-of-the-art performance with over 3X reduction in transmitted pixels.

\end{abstract}

\vspace{-15pt}
\section{Introduction}
While edge camera IoT devices, which connect the physical world to the cloud, are revolutionizing data gathering in many consumer and business applications, their limited resources place a significant burden on their service life and operation, warranting cost-aware inference methods. %that achieve state-of-the-art accuracy of a fully trained deep neural network with minimal resource utilization. 
For low-powered IoT devices, transmit energy dominates all other forms of battery usage \cite{halgamuge2009estimation,bsn}, and so, we are justified in considering the number of edge-cloud interactions as a surrogate for battery usage. Our goal is to minimize pixels transmitted while simultaneously ensuring accuracy that is comparable to a fully trained state-of-art deep neural network\footnote{The minimum amount of pixels for a standard Inception-V3 input size is about $299\times299$ to achieve good accuracy. The pictures captured by modern high resolution cameras are often larger than this size.}, which has access to the entire image. Our prototypical setup is an edge-device that sparingly transmits image regions to a cloud server equipped with abundant computational resources (e.g. a Inception-V3 network~\cite{szegedy2016rethinking}), to interpret received inputs\footnote{This edge-cloud setup is a conventional model in  IoTs (see ~\cite{perera2015emerging}).}. 

Fine-grained classification poses fundamental challenges in the IoT setup and highlights a fundamental dilemma between accuracy and cost: {\it On the one hand an instance cannot be classified accurately, unless the cloud sees the most discriminative parts. On the other hand an edge device can neither transmit the entire image due to bandwidth/power constraints, nor can it locally identify those parts due to lack of computational resources.} 

\begin{figure*}[t]
\centering
\includegraphics[width=0.95\textwidth]{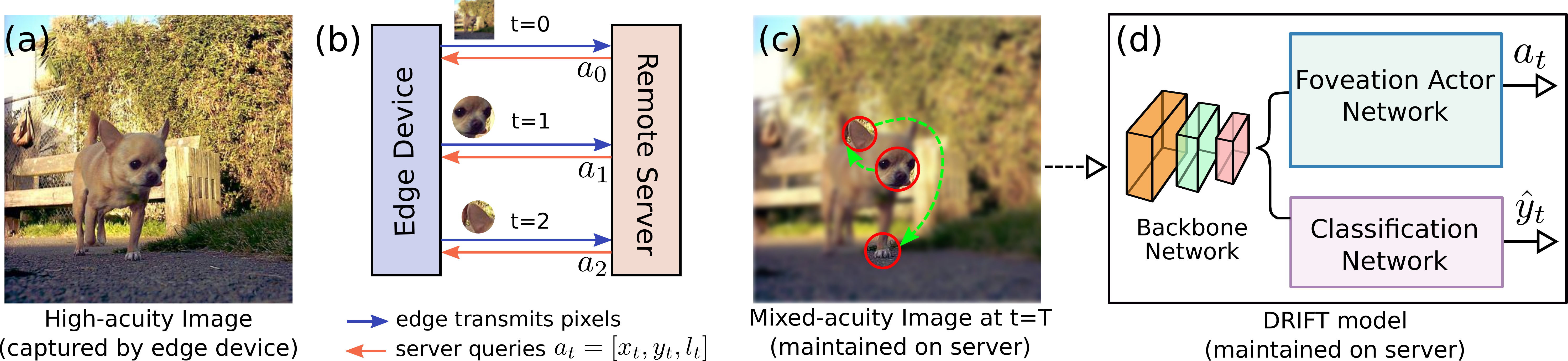}
\caption{(a) The original high-acuity image captured by a low power edge device;
(b) The IoT scenario: At $t=0$ the edge device only transmits a thumbnail with extremely low resolution, e.g. $30\times30$, to the server.  The server then calculates the next fixation point (location and size) to query more high-acuity pixels from the edge device;
(c) The mixed-acuity image which includes all the received pixels so far at $t=T$;
(d) The proposed DRIFT model, described in Sec.\ref{sec:method}. }
\vspace{-5pt}
\label{fig:fig2}
\end{figure*}

In this context we are compelled to adopt a novel interactive edge-cloud model. Our approach is a novel deep reinforcement learning-based foveation model, DRIFT, that sequentially generates and recognizes mixed-acuity images. Instead of transmitting the full details all at once, the edge device first transmits a preliminary coarse thumbnail, e.g. $30\times30$, to the server, which then actively but sparingly interacts with the edge device to seek image regions of value (see Fig.\ref{fig:fig2}), based only on the past received inputs. To summarize, our method (1) operates on mixed-acuity inputs whose resolution varies across the image, and more importantly,
(2) actively and sequentially determines which image regions should be perceived with more visual details based on a low-acuity input, without access to the class label, a process we humans perform naturally\footnote{When we humans view a novel scene we do not perceive its full complexity at once, but rather, we foveate~\cite{mcconkie1975span}. In doing so, our brain processes information from high-acuity foveal regions and the coarser-resolution periphery. 
%The resulting process is additive: 
% Starting with an initial coarse view, we `fill in' further details via fixations (see Fig.\ref{fig:fig2}).
The ability to process mixed-acuity inputs and actively `choose' where to fixate significantly increases our efficiency, particularly because we do not need to observe every detail before recognizing a scene.}. %We propose a novel Deep ReInforcement FoveaTion (DRIFT) model inspired by humans who sequentially fixate with a foveated field-of-view 
%While modeling foveation in biologically-plausible ways is challenging, we are motivated  by the top-down process that  (1) infers fixation points from low-acuity images where most of the contents are blurred; and (2) sequentially refines the next fixation based on the newly received high-acuity image contents. 
%  
%
%identifying the important discriminative parts by the cloud, agnostic to the full image, may require more edge-cloud interactions, since the edge device has little computational resources at hand.} 
%
While existing trained deep network architectures are an option, they are fundamentally ill-suited for this distributed task, and their variants, namely, attention-based methods require the entire high-acuity image to be available up-front.

Our DRIFT model consists of three neural networks: (1) a backbone CNN to extract visual features from low/mixed-acuity input images; (2) a foveation actor network to generate a sequence of fixation actions; 
%i.e. the location of each fixation point and the size of the high-acuity region; and 
and (3) an image classification network to predict the final class label.
%To train the foveation actor network, 
% we use the classification network to decide whether a foveated image contains sufficient information for the input to be correctly classified. To achieve this, 
We propose a novel reinforcement learning (RL) reward to guide the training so that the model fixates on regions that lead to high accuracy, while limiting the total transmitted high-acuity pixels. 
Given a low-acuity image transmitted from the remote edge device, a cloud server with DRIFT is able to predict locations of the most discriminative visual cues and actively query and incorporate such visual details by further interacting with the edge device.

Since the space for a fixation action (location and size) is large, discretizing/enumerating this space would be intractable in practice. Therefore, we propose solving in continuous space and train a foveation policy with a novel Deep Deterministic Policy Gradient by Conditioned Critic with Coaching (DDPGC3) algorithm. Compared to the original DDPG algorithm~\cite{lillicrap2017continuous}, several modifications are made:
First, DDPG trains a critic to approximate an action-value function~\cite{watkins1992q} to evaluate the learned policy, and uses the evaluation results to guide reinforcement learning. While this action-value function is globally shared among all state-action pairs in \cite{lillicrap2017continuous}, we found this global function is difficult to approximate in our foveation problem. Instead, we propose training the critic to approximate a conditioned state-value function that is uniquely defined on every RL episode and more easy to approximate. Second, the actor network parameters in \cite{lillicrap2017continuous} are updated completely based on the critic's evaluation. However, at the early training stage a deficient critic can easily misguide the updates. Consequently, we propose updating the actor network by coaching~\cite{he2012imitation}, i.e. by both the critic's policy evaluation as well as the actions generated by a heuristic oracle. We observe that our improvements on DDPG stabilizes the training procedure.

\noindent \textbf{Contributions}: 
\textbf{(1)} An active image acuity exploration model, DRIFT, is proposed, for IoT applications which require efficient data transmission.  DRIFT is able to sequentially infer fixation points from low-acuity images and classify based on mixed-acuity inputs;
\textbf{(2)} A novel reward function is introduced so that the proposed DRIFT model can be trained with weak image-level class labels instead of more fine-grained labels on locations and sizes;
\textbf{(3)} We propose training a foveation actor network via (I) a conditioned critic that approximates a unique state-value function conditioned on every input image, and (II) a coaching mechanism that combines the critic's evaluation with a heuristic oracle; 
\textbf{(4)} Experiments on five fine-grained classification datasets 
% CUB-200-2011~\cite{wah2011caltech}, Stanford Cars~\cite{KrauseStarkDengFei-Fei_3DRR2013}, Stanford Dogs~\cite{imagenet_cvpr09}, Aircrafts~\cite{maji13fine-grained}, and Food-101~\cite{bossard14}. 
show that DRIFT achieves competitive performance with substantially fewer pixels compared to standard deep CNN models. %thanks to its foveation mechanism. 
\textbf{(5)} Furthermore, since DRIFT discovers discriminative visual features, it can also be used to generate hard attention which boosts standard classification performance for existing deep CNN models.
Finally, although we demonstrate the proposed model's effectiveness in classification, DRIFT is a general active image acuity exploration solution which can be applied to other domains.

% \begin{figure*}[h]
% \centering
% \includegraphics[width=0.9\textwidth]{images/state.pdf}
% \caption{Illustration on state components.}
% \label{fig:state}
% \end{figure*}

\section{Related Work}
%\todo[inline]{vxa: I would begin by emphasizing that 
While there are a number of works that deal with edge-computing, wireless sensor networks (see ~\cite{yang2008,bsn,perera2015emerging}) and resource constrained learning (see ~\cite{NIPS2017_7058,pmlrv89zhu19d,bolukbasi2017resource,pmlrv54hanawal17a,trapeznikov2013supervised,wang2014lp}), the focus of these works tend to be one-shot, and interaction between cloud and edge device is not considered. DRIFT is the first to propose a sequential decision making process, whereby a cloud with computational resource interacts with a bandwidth limited edge device. This interaction is necessitated by fine-grained classification problem.% in the context of limited edge resources. 
%DRIFT is the first attempt targeted

Recently, various attention models~\cite{xu2015show,fu2017look,li2018tell,shih2016look,mathe2016reinforcement} have been proposed to enable CNNs to attend to specific image regions for multiple vision tasks. Nonetheless, these methods are optimized for foveated interaction leading to inefficient data transmission. %still intrinsically different and less efficient in data transmission than a true foveation system. %
In particular, they still operate on single high-acuity level inputs, and cannot sequentially infer attentions from low/mixed-acuity inputs. Consequently, all image details have to be revealed as inputs, a priori, instead of accumulation of a sequence of fixations.

Our research is related to foveation. Deng \etal~\cite{deng2013fine} 
found that humans were able to correctly recognize an object by only revealing few high-acuity regions (referred to as fixations here) on a heavily blurred image. They thus propose crowd-sourcing to collect such location annotations and training detectors on these discriminative features to boost classification accuracy. Matzen \etal~\cite{matzen2015bubblenet} extended this concept and proposed an automatic but brute-force approach by initializing hundreds of random fixations per image, and iteratively optimizing to adjust each fixation based on the classification scores of their corresponding foveated images. These brute-force approaches require too many fixation regions to be transmitted per image, and are thus incompatible with IoT setup.

Different from~\cite{deng2013fine,matzen2015bubblenet}, which take low-acuity inputs, Almeida \etal~\cite{almeida2017deep} and Recasens \etal~\cite{recasens2018learning} proposed generating attention maps from images with standard input sizes, to either down-sample backgrounds~\cite{almeida2017deep} or up-sample foregrounds~\cite{recasens2018learning}. The approach of generating attention maps falls into a broader family of attention models, which has been predominantly applied to image classification~\cite{mnih2014recurrent,ba2014multiple,zhao2017diversified,fu2017look,zheng2017learning}, segmentation~\cite{li2018tell}, visual question answering~\cite{shih2016look,liang2018focal}, detection~\cite{zhang2018progressive}, image captioning~\cite{xu2015show}, and so forth. 
As is the case for deep CNNs, these attention models require the full high-acuity images to be transmitted, and their performance degrades significantly on low-acuity inputs (see Sec.~\ref{sec:experiment}).
In contrast our paper focuses on automatically inferring fixations from extremely low-acuity inputs (e.g. $30 \times 30$). We take a sequential and additive approach: The proposed DRIFT model is able to accumulate knowledge, recursively refine its fixations, and finally produce fixation locations that are optimized for classification accuracy as well as data efficiency. %Benefiting from a reinforcement learning formulation, %
DRIFT learns to avoid exhaustive search, and thus is 
superior to the brute-force approach in \cite{matzen2015bubblenet}.

In our RL formulation, fixations are modeled as a sequence of actions generated by a foveation actor model, which is similar in spirit to a few RL-based object detection works, e.g.~ \cite{mathe2016reinforcement,pirinen2018deep,jie2016tree,bueno2017hierarchical,caicedo2015active}. However, our work is significantly different in that: 
(1) the proposed DRIFT learns fixation actions without any supervision on object locations, therefore it is more scalable to large scale data;
(2) our action space is infinite, whereas in \cite{pirinen2018deep,jie2016tree,bueno2017hierarchical,caicedo2015active} the actions can only be chosen within a restricted list, which limits diversity of model outputs; and
(3) our low-acuity input uses much less information compared to the full-resolution input images in these detection methods. %Therefore, the foveation problem is more challenging in all dimensions of input, output and supervision.

\vspace{-0.1cm}
\section{Methodology}
\label{sec:method}
\vspace{-0.1cm}
\subsection{Foveation for Internet-of-Things}
\label{sec:foveation_for_classification}
%Let us formally define our setting. To narrow down the horizon of this study, 
While our method is general, for concreteness we consider foveation within the context of image classification. %Note that the general idea can be applied to other domains.
We assume there are two types of representations for any scene/object: a low-acuity image $I_{low}$ with limited visual details, and a high-acuity image $I_{high}$. For example, $I_{high}$ could have standard Inception-V3 input size $299\times299$,
and $I_{low}$ is a down-sampled version with size $30\times30$.
A low-battery edge device could directly transmit all the details on $I_{high}$ to a cloud server which leads to high classification accuracy, but to transmit $I_{high}$ is expensive.

Our {\bf foveation for IoT pipeline} is then defined as: 
(1) An edge device transmit $I_{low}$ as an initial input to a cloud server. 
(2) A foveation model on the server infers a fixation point, which defines the coordinates and radius of a  small circular image region. The coordinates and radius are then sent back to the edge device. 
(3) The edge device only transmits new high-acuity contents on $I_{high}$ as specified by the fixation point.
(4) The server incorporates the newly received high-acuity pixels to $I_{low}$, updates its posterior, and requests new coordinates if not confident.
Intuitively, a good foveation model should reach a balance between {\it accuracy} and {\it transmission efficiency}, i.e. it should fixate on the most discriminative image regions so that good classification results can be achieved, while keeping the overall transmitted high-acuity pixels at a very low amount.

% The high-acuity content at this fixation point on $I_{low}$ is revealed, meaning that it is replaced by the corresponding region from $I_{high}$. The resulting foveated image, $I_{fovea}$, has low-acuity content overall but high-acuity content only at fixation point(s). 
% $I_{fovea}$ can be used as input to repeat step (1) and generate the next fixation point, or directly used for classification after a few iterations. 

We view this foveation pipeline as an instance of Markov Decision Problem (MDP), and adopt reinforcement learning (RL) for training. We employ RL to train the foveation model because
the optimal foveation policy should not be learned with any explicit supervision other than the objective of optimizing transmission efficiency and classification accuracy. It is thus difficult to define such a loss via standard supervised learning, but in RL, this training objective can be easily reflected by a reward function.
% (1) The location and size of each fixation point has to be decided sequentially, which RL naturally lends itself to; 
% (2) The optimal foveation policy should not be learned with any explicit supervision other than the single objective of optimizing classification accuracy. It is thus difficult to define such a training loss for traditional supervised learning, but in RL, this training objective can be easily reflected by a reward function;
% (3) Revealing high-acuity image regions at fixation points is a non-differentiable procedure impeding back-propagation, which can be mitigated by policy gradients algorithms~\cite{silver2014deterministic,gu2017q,gu2017interpolated} in RL; 
% (4) The sequential fixation points are dependent and there exists an exponentially large number of ordering and combinations. For example, in bird species recognition, a foveation model could start from the beaks for some instances, but wings for others. Exploration-exploitation strategies in RL can be applied so that such interdependencies are better captured by the foveation model as the large action space is being explored. 
%

\subsection{Markov Decision Process Formulation}
We consider foveation for IoT as a sequential decision making problem, where the foveation model interacts with a dynamic environment $E$ at discrete timesteps.
At each timestep $t$, the foveation model receives an observation state $s_t$, takes an action $a_t$, and receives a scalar reward $r_t=r(s_t, a_t)$. This MDP process can be formally modeled by: action space $\mathcal{A}$, state space $\mathcal{S}$, transition dynamics from $s_t$ to $s_{t+1}$ after receiving $a_t$, and reward function $r(s_t, a_t)$.  The foveation model implements a policy function $\pi$, which maps states to a distribution over actions: $\mathcal{S} \rightarrow \mathcal{P}(\mathcal{A})$. The return at timestep $t$ is defined as the sum of discounted future rewards $R_t = \sum_{i=t}^\infty\gamma^{(i-t)}r(s_i, a_i)$ with a discount factor $\gamma \in [0,1]$. Note that the return $R_t$ depends on the actions taken, and thus depends on the policy $\pi$. The goal of RL is to find the best policy which maximizes the expected return $\mathbb{E}_{r_{t}, s_{t} \sim E, a_{t} \sim \pi} [\sum_{t=0}^\infty\gamma^{(t)}r(s_t, a_t)]$. Next we explain in detail each component of our MDP.

\begin{figure*}[t]
\centering
\includegraphics[width=0.8\textwidth]{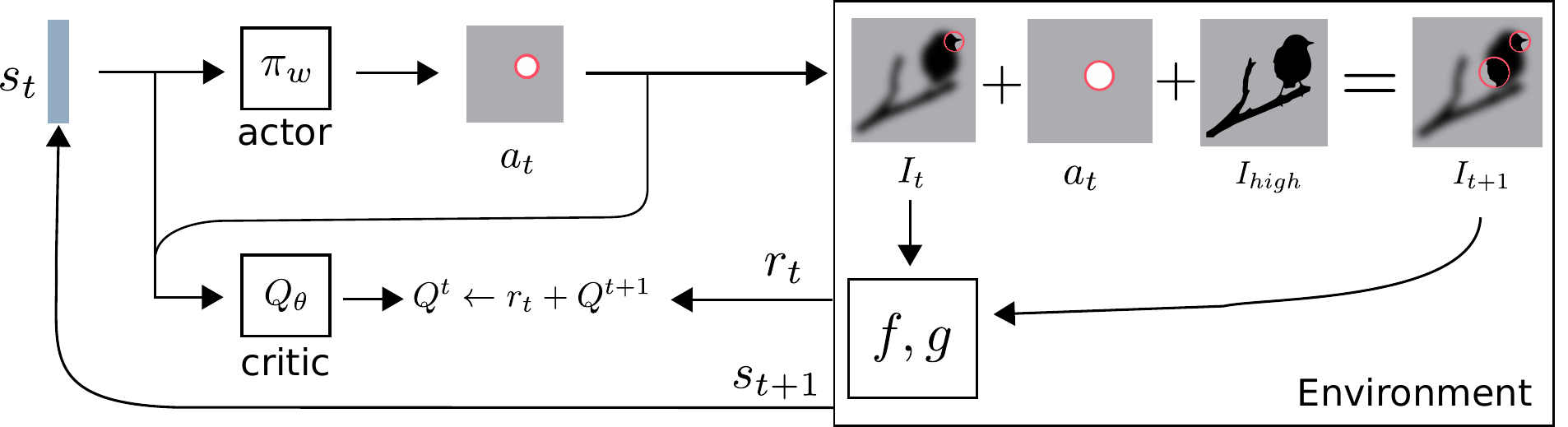}
\caption{Illustration on our RL pipeline. The critic $Q_\theta$ is explained in Sec.\ref{sec:DDPGC3}.}
\label{fig:pipeline}
\vspace{-5pt}
\end{figure*}

\noindent \textbf{Episode}: To mimic the IoT scenario,  we take images of the standard input size (e.g. $299\times299$ for Inception-V3) as $I_{high}$, and down-sample it as $I_{low}$ (e.g. $30\times 30$). $I_{low}$ is thus left with very limited visual details (see Fig.\ref{fig:bboxes}).
At $t = 0$, the edge device transmits $I_{low}$. On receiving $I_{low}$, the server-side environment $E$ interpolates it back to input size and uses it as the initial input for the foveation model ($I_t = \Psi(I_{low})\footnote{$\Psi(\cdot)$ refers to linear interpolation.}, t = 0$); 
At each time step $t \in \{0, 1, \cdots, T-1\}$ where $T$ is a pre-defined episode length, the foveation model predicts a fixation action $a_t$ based on $I_t$, where $a_t$ specifies the location and size of a small circular image region; On receiving $a_t$, the edge device only needs to further transmit new pixels specified by $a_t$ in $I_{high}$.
Environment $E$ then renders $I_{t+1}$ to the foveation model by replacing the low-acuity content within the region on $I_t$ specified by $a_t$ with the newly received high-acuity pixels.
Finally, at the end of each episode, the foveation model predicts a class label based on all accumulated pixels $I_T$. %which contains all the pixels transmitted so far.

\noindent \textbf{Action Space}:  The fixation action $a$ generated by the foveation model includes the predicted spatial location and size of a small circular image region. Specifically,
\begin{equation}
    a = (x, y, l), \quad x, y, l \in [-1, 1],
\end{equation}
where $(x, y)$ refer to the horizontal and vertical coordinates of the fixation center, and $l$ the radius. 
To facilitate training, the actions are normalized to $[-1, 1]$ rather than in real pixels. Suppose the original image size is $(h, w)$, and the smallest and largest fixation point radius are predefined by $b_1$ and $b_2$. With action $a = (x, y, l)$, the real location and size of a fixation point can be obtained by $(\frac{(1+x)}{2}w, \frac{(1+y)}{2}h, b_1 + \frac{(1+l)}{2} (b_2-b_1))$.

\noindent \textbf{State Space}: As illustrated by Fig.\ref{fig:fig2}, we have a backbone network $f$ and a classification network $g$, where $f$ extracts visual features for any given input image, and $g$ maps the extracted features to classification predictions. At time step $t$, the state $s_t$ of the observation $I_t$ is given by: %(See Fig.\ref{fig:state}.)
\begin{equation}
s_t = [f(I_t), f(I_{t-1}), f(I_{t}^{local}), h_t], 
\end{equation}
where $f(I_t)$ and $f(I_{t-1})$ are the feature vectors of the current and last step observations; $f(I_{t}^{local})$ is the feature vector of the local image patch (resized to input size) on $I_{t}$ around the the newest fixation point $a_{t-1}$; $h_t \in \mathbb{R}^{dim(a)\times T}$ is an action history vector, represented by the concatenation of the past actions $[a_0, a_1, \cdots, a_{t-1}, \mathbb{O}]$, with future actions padded by zeros. 

\noindent \textbf{Initial State}: At $t=0$, the state $s_0$ is initialized by $[f(I_0), \mathbb{O}]$, with $f(I_{t-1}), f(I_{t}^{local}), h_t$ padded by zeros.

\subsection{Dense Reward by Relative Comparison} 
\noindent Our goal is to achieve high accuracy with the foveated image at $t=T$, with minimum high-acuity content explored by its fixation actions. For example (Fig.\ref{fig:fig2}), to recognize Chihuahua, good fixations should be focused on discriminative characteristics of the Chihuahua, such as its face and ears. A naive strategy is to check at the end of each episode whether $I_T$ can be correctly classified by $g$. However, this type of reward provides only episode-level feedback which is sparse and stale, and thus difficult to associate with single actions (credit assignment problem~\cite{pathak2017curiosity}).

As a solution, we propose a dense reward function defined at each time step $t$. Specifically, given action $a_t$, the observation changes from $I_t$ to $I_{t+1}$, with the high-acuity region specified by $a_t$ revealed. Given the ground truth label $y$ for current episode, we calculate two cross-entropy losses $\ell_t^1 = XE(g(f(I_t)), y) $ and $\ell_t^2 = XE(g(f(I_{t+1})), y)$. Intuitively, a good fixation $a_t$ should increase the classification model's confidence and make $\ell_t^2 < \ell_t^1$. The \textbf{accuracy reward} is thus given by a relative comparison:
\begin{equation}
    r_t^a = \ell_t^1 - \ell_t^2
\end{equation}
In addition, we want to restrict the overall high-acuity content and prevent brute-force fixations. Let $p_t$ denote the overall amount of high-acuity pixels revealed at $t$, $thr$ a pre-defined threshold, $\mathbb{I}(\cdot)$ the indicator function, our \textbf{transmission efficiency reward} is:
\begin{equation}
    r_t^e = - \mathbb{I}(t = T,  p_t > thr)
\end{equation}
Reward $r_t$ is then defined by the sum: $r_t = r_t^a + \lambda r_t^e$, where $\lambda$ is a hyperparameter controlling the trade-off between accuracy and transmission efficiency.

\subsection{DDPG by Conditioned Critic with Coaching}
\label{sec:DDPGC3}
% We next describe the proposed RL algorithm for training the foveation model, termed as Deep Deterministic Policy Gradient by Conditioned Critic with Coaching (DDPGC3), started by a brief introduction on the original DDPG~\cite{lillicrap2017continuous}.

\noindent \textbf{Deep Deterministic Policy Gradient} Recently proposed by Lillicrap \etal \cite{lillicrap2017continuous}, the DDPG algorithm trains deep neural networks to learn policies in high-dimensional, continuous action spaces, and thus is suitable for our problem. The key insight of DDPG is to apply an actor-critic setup~\cite{silver2014deterministic}.
Specifically, we assume the policy $\pi$ is modeled by an actor network parameterized by $w$, which outputs a continuous deterministic policy $a = \pi_w(s)$. 
To optimize the policy, it takes a typical policy evaluation and improvement scheme.
Policy evaluation uses state-value function $Q(s_t, a_t)$ to evaluate the current policy's expected return, where $Q(s_t, a_t) = \mathbb{E}_{r_{i\geq t}, s_{i>t} \sim E, a_{i>t} \sim \pi}(R_t | s_t, a_t)$. Here the state-value function is approximated by a critic network parameterized by $\theta$, denoted as $Q_\theta$, which only serves to train the actor network and is discarded during testing.
Policy improvement uses the critic's estimation to improve the current policy model so that better $Q(s_t, a_t)$ can be reached.
%\noindent

Formally, the critic network is trained to optimize the temporal-difference (TD) term of the Bellman equation:
\begin{align}
    J_\theta &= \min_\theta \mathbb{E}_{s_t, a_t, r_t \sim \beta} [(Q_\theta(s_t, a_t) - q_t)^2], 
\label{eq:critic}    
\end{align}
where $q_t = r_t + \gamma Q_{\theta'}(s_{t+1}, a_{t+1})$, $\beta$ is the distribution of off-policy $(s_t, a_t, r_t, s_{t+1}, a_{t+1})$ samples stored in a replay buffer, $Q_{\theta'}$ is a separate target network used to generate TD targets $q_t$. The weights of the target
network are updated by having them slowly track the learned networks: $\theta'=\tau \theta+(1-\tau) \theta'$ with $\tau \ll 1$. Both the replay buffer and target network are originally introduced in \cite{mnih2015human} to de-correlate training samples and stabilize the training process.

In \cite{lillicrap2017continuous} the objective for training the actor network is simply to maximize the critic's estimation:
\vspace{-0.05in}
\begin{equation}
    J_w = \max_w \mathbb{E}_{s_t, a_t, r_t \sim \beta} [Q_\theta(s_t, \pi_w(s_t))].
\label{eq:actor}    
\vspace{-0.05in}
\end{equation}
\noindent \textbf{Conditioned Critic with Coaching} We observe that DDPG fails to train a good foveation actor. Our analysis follows:

First, the global state-value function $Q(s_t, a_t)$ is too difficult to be approximated by the critic network $Q_\theta(s_t, a_t)$. Intuitively, given $(s_t, a_t)$, in the original formulation (Eq.\ref{eq:critic}), the critic network is expected to estimate $R_t$, which reflects the reward $r_t$.
$r_t$ depends on the ground-truth label $y$ and $g$'s prediction $\hat{y}$, while $\hat{y}$ further depends on the high-acuity region specified by $a_t$.
% (1) the high-acuity region specified by $a_t$ on $I_{high}$; (2) the classification network $g$'s new prediction $\hat{y}$; (3) the ground truth label $y$; and (4) the reward $r_t$ which depends on $(y, \hat{y})$. 
Since the critic does not have access to any of $\{I_{high}, g, y, r_t\}$ by definition, the estimation of $R_t$ is difficult. Observing this issue, we propose training a \textbf{conditioned critic} which approximates a unique state-value function defined for each episode $k$, $Q(s_t, a_t | C^k)$, where $C^k = [f(I_{high}^k), y^k]$ is the condition, $I_{high}^k$ and $y^k$ referring to the high-acuity image and ground-truth label for the $k$-th episode. 
%Compared to the global state-value function, the conditioned version is much easier to be learned.
By substituting the conditioned critic in Eq.\ref{eq:critic}; letting $q_t = r_t + \gamma Q_{\theta'}(s_{t+1}, a_{t+1} | C^k)$, and $\psi$ as the distribution of episodes, we have a new objective for critic training:
\begin{align}
    J_\theta &= \min_\theta \mathbb{E}_{s_t, a_t, r_t \sim \beta, C^k \sim \psi} [(Q_\theta(s_t, a_t | C^k) - q_t)^2], 
\label{eq:critic_new}    
\end{align}
%where $q_t = r_t + \gamma Q_{\theta'}(s_{t+1}, a_{t+1} | C^k)$, and $\psi$ is the distribution of episodes.
%
Second, in Eq.~\ref{eq:actor} the actor network's optimization is solely based on the critic's estimation. Since the critic network parameter $\theta$ is randomly initialized, the critic's estimation is initially a random guess. This significantly slows down the training process and impedes convergence of actor training. To solve this problem, we leverage the idea of coaching~\cite{he2012imitation} and introduce a low-cost heuristic oracle\footnote{We use the term heuristic since a true oracle is impossible to realize without exhaustively searching over the large action space.}, as in  ~\cite{zhou2018weakly}, which provides a policy better than random guessing, and can be also used to guide early stage actor training. The actor training by \textbf{coaching} objective is defined as:
\vspace{-0.1in}
\begin{align} \nonumber
%\begin{split}
    J_w = \max_w \mathbb{E}_{s_t, a_t, r_t, a_t' \sim \beta, C^k \sim \psi} & [(1-\epsilon)Q_\theta(s_t, \pi_w(s_t) | C^k) \\ \label{eq:actor_new}
    &- \epsilon|\pi_w(s_t) - a_t'|^2],
%\end{split}
\end{align}
where $a_t'$ is the action taken by the heuristic oracle given $s_t$,  and $\epsilon$ is a exponentially decreasing factor with respect to the training progress. We refer to this actor-critic RL training strategy with Eq.~\ref{eq:critic_new}-\ref{eq:actor_new} as DDPG by Conditioned Critic with Coaching (DDPGC3).
\begin{figure}[t]
\centering
\includegraphics[width=0.35\textwidth]{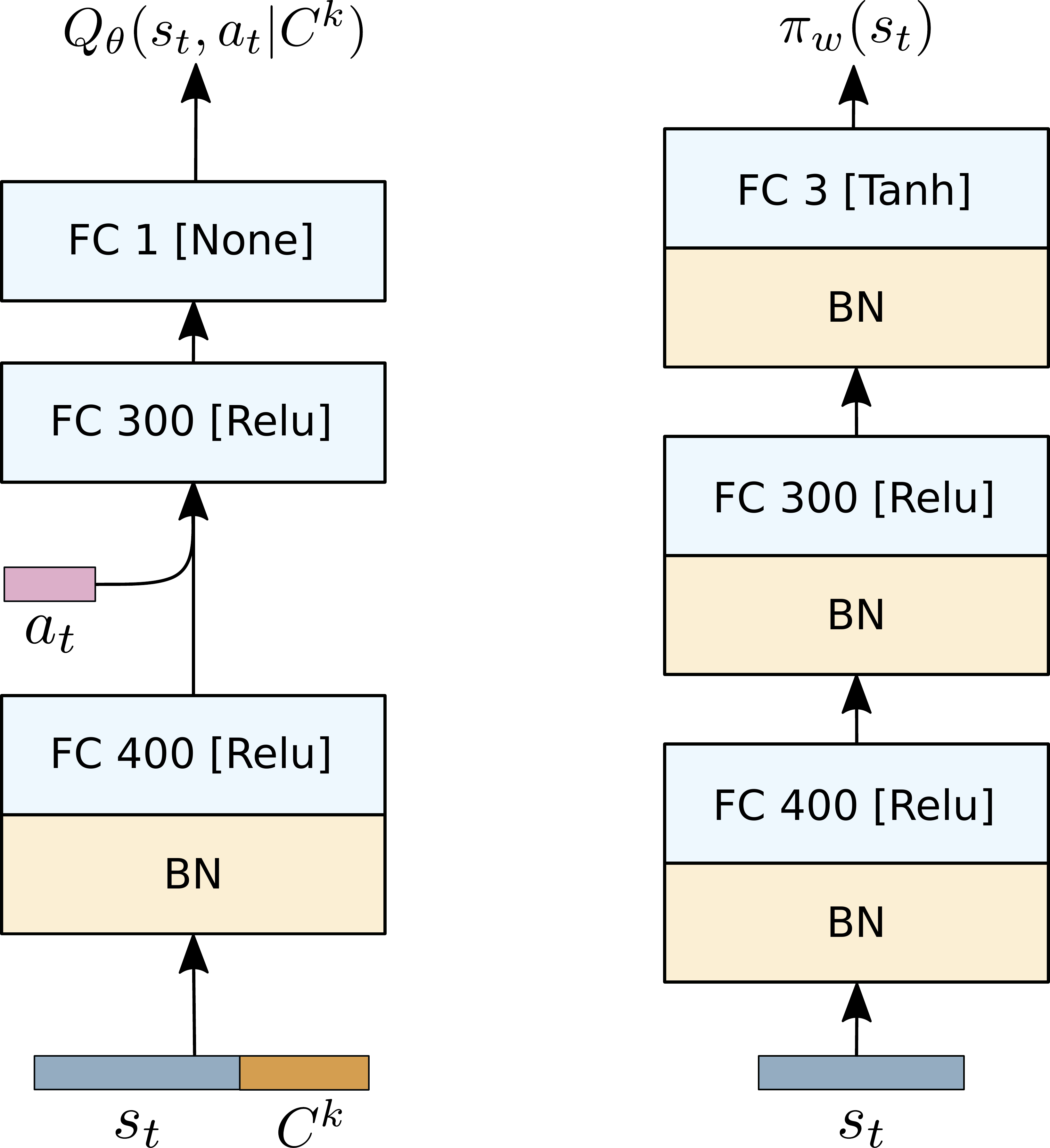}
\caption{Our critic (left) and actor (right) network architecture. FC: fully-connected layer. BN: batch normalization layer. Numbers: the amount of neurons. Brackets: activation functions.}
\label{fig:network}
\end{figure}
\begin{table*}[t]
\begin{minipage}[t]{0.6\textwidth}
\centering
\scalebox{0.85}{
\renewcommand{\arraystretch}{1.1}
\setlength{\tabcolsep}{0.15cm}
\begin{tabular}{|l | c c | c c | c c | c c | c c |}
\whline
datasets & \multicolumn{2}{c|}{CUB} & \multicolumn{2}{c|}{Cars} & \multicolumn{2}{c|}{Dogs} & \multicolumn{2}{c|}{Aircrafts} & \multicolumn{2}{c|}{Food101} \\
(\%) & acc & pix & acc &  pix & acc &  pix & acc &  pix  & acc &  pix\\
\hline
\hline
Random & 30.1 & 15.0 & 44.1 & 15.0 & 33.2 & 15.0 & 38.4 & 15.0 & 40.8 & 15.0\\
\rowcolor{Gray} 
Center & 59.3 & 15.0 & 55.3 & 15.0 & 58.4 & 15.0 & 69.6 & 15.0 & 42.3 & 15.0\\
Saliency  &  37.5 & 15.5 & 29.7 & 14.8 & 39.8 & 16.1 & 33.6 & 15.4 & 28.1 & 12.2\\
\rowcolor{Gray} 
Attention & 44.4 & 15.6 & 51.5 & 14.9 & 46.4 & 14.8 & 63.6 & 14.9 & 35.5 & 14.8 \\
BubbleNet & 65.5 & - & -  & - & - & - & - & - & 56.1 & -  \\
% resize & 66.2 & 10.1 & 77.8 & 11.5 & 64.3 & 14.1 & 66.0 & 14.4 & & \\
\rowcolor{Gray} 
DRIFT  & \bf 74.4 & \bf 10.1 & \bf 82.8 & \bf 11.5 & \bf 71.6 & \bf 14.1 & \bf 86.7 & \bf 14.4 & \bf 75.5 & \bf 11.4\\
\hline
\hline
 $I_{low}$ & 13.9 & 1.0 & 7.8 & 1.0 & 17.1 & 1.0 & 6.6 & 1.0 & 8.9 & 1.0\\
\rowcolor{Gray} 
$I_{high}$ & 81.6 & 100.0 & 91.2 & 100.0 & 81.8 & 100.0 & 87.2 & 100.0 & 85.0 & 100.0 \\
  DRIFT$_E$ & 80.1 & 32.6 & 88.5 & 33.6 & 78.0 & 35.6 & 88.0 & 35.8 & 81.9 & 33.6\\
\whline
\end{tabular}}
\caption{IoT classification setting. DRIFT outperforms other foveation methods, while requiring substantially fewer pixels to be transmitted. \label{table:foveat}}
\vspace{-5pt}
\end{minipage}
\begin{minipage}[t]{0.4\textwidth}
\scalebox{0.9}{
\renewcommand{\arraystretch}{1.05}
\setlength{\tabcolsep}{0.15cm}
\begin{tabular}{|l | c  c  c  c  c|}
\whline
datasets (\%) & CUB & Cars & Dogs & Air & Food \\
\hline
\hline
Bilinear~\cite{lin2015bilinear} & 84.1 & 91.3 & - & 84.1 & - \\
\rowcolor{Gray}RA-CNN~\cite{fu2017look} & 85.3 & 92.5 & 87.3 & 88.2 & - \\
FCAN~\cite{liu2016fully} & 84.3 & 91.5 & \bf 88.9 & - & 86.3\\
\rowcolor{Gray}GP~\cite{wei2018grassmann} & 85.8 & 92.8 & - & 89.8 & 85.7 \\
MAMC~\cite{sun2018multi} & 86.2 & 92.8 & 84.8 & - & - \\
\rowcolor{Gray}DFL-CNN~\cite{wang2018learning} & \bf 87.4 & 93.1 & - & 91.7 & - \\
\hline
Inception-V3 &  81.6 & 91.2 & 81.8 & 87.2 & 85.0 \\
\rowcolor{Gray}DRIFT$_I$  &  83.7 & 92.2 & 82.9 & 90.7 & 86.6\\
\hline 
 ResNet-50 & 83.2   & 92.2  & 85.7  & 89.8  & 85.8  \\
\rowcolor{Gray} DRIFT$_R$  & 86.2  & \bf 93.6  &  87.3 & \bf 91.7  & \bf 88.6  \\
\whline 
\end{tabular}}
%\vspace{-0.3cm}
\caption[caption]{Standard classification setting. DRIFT consistently improves baselines' performance. \footnotemark \label{table:normal}}
\vspace{-5pt}
\end{minipage}
\end{table*}
Using the final feature map prior to spatial pooling in $f$, we perform a $1\times 1$ convolution with the ground-truth class's classifier to get a response map $m$ (for Inception-V3: feature map is shaped $8\times 8\times 2048$ and $m$ is $8 \times 8$). We then sample a location $(x', y')$ based on $m$'s values, randomly sample a radius $l' \in [-1 , 1]$, and use $(x', y', l')$ to construct $a_t'$. Even though our naive $a_t'$ only provides a coarse clue on the classifier's response over the low-acuity observation $I_t$, it still helps to speed up and stabilize the actor training by significantly saving efforts spent on random explorations caused by the deficient critic during early training. The oracle, which has access to the GT label, is only used during training and discarded when testing.

\subsection{Implementation, Training and Deployment}
\noindent \textbf{Implementation} We implemented our model using Tensorflow~\cite{abadi2016tensorflow}. For the backbone $f$ and classification network $g$, we adopted the Inception-V3 architecture~\cite{szegedy2016rethinking}, i.e. $f$ outputs a $2048$-d feature vector, and $g$ is a fully-connected layer followed by a softmax. 
The architectures for our actor network $\pi_w$ and critic network $Q_\theta$ are illustrated in Fig.\ref{fig:network}.
For a given backbone with a default input size, e.g. $299 \times 299$ for Inception-V3, we define $I_{high}$ to be the standard input image, and generate $I_{low}$ by down-sampling $I_{high}$ to $30\times 30$ (only retaining $1\%$ pixels).
We set the smallest and largest fixation radius $b_1, b_2$ to 15 and 75, episode length $T$ to 5, data-efficiency reward trade-off $\lambda$ to $5.0$, threshold $thr$ to $25\%$ of $I_{high}$ pixels, discount factor $\gamma$ to $0.9$, target network update rate $\tau$ to $1e^{-4}$. As in~\cite{lillicrap2017continuous}, we add Ornstein-Uhlenbeck noise~\cite{uhlenbeck1930theory} to our actor policy for exploration.

\noindent \textbf{Training} We first pretrain $g \circ f$ on $I_{high}$ with a standard classification loss. Then we train $\pi_w$ and $Q_\theta$ by the proposed DDPGC3 algorithm (Sec.~\ref{sec:DDPGC3}) for 60 epochs, with a SGD optimizer, a batch size of 32, and a fixed learning rate of $1e^{-4}$. 
For the beginning 50 epochs we freeze $f$ and $g$, and then in the remaining epochs $f$ and $g$ are updated by a standard classification loss with the foveated images $I_T$ as input. The size of experience replay buffer $\beta$ for RL was 50,000. The decreasing factor $\epsilon$ for coaching in Eq.~\ref{eq:actor_new} is set to $0.7$ initially and decays $0.96$ every $1000$ training updates. During training, $(s_t, a_t, a_t', r_t, s_{t+1}, a_{t+1})$ samples are first pushed into the replay buffer, and then randomly sampled to update the actor and critic.

\noindent \textbf{Deployment} The critic network is deleted after training. 
The part $[f, \pi_w, g]$ is maintained at a cloud server.
Once a low-acuity image is received, 
$\pi_w \circ f$ networks can be used to generate sequential fixation points, and $g \circ f$ to classify the resulting foveated images. 
%Note that, in this paper we focus on foveation with {\it transmission efficiency} while assuming a sufficient computational power at the server side. 
Thus, both $\pi_w$ and $g$ take features generated by a shared backbone $f$. In real-world deployments where {\it server computation efficiency} is required, we see that we place no limitation on $\pi_w$ to use cheap features while $g$ may use expensive features so that a balance between data and computation efficiency might be reached.

\footnotetext{In both Table~\ref{table:foveat} and~\ref{table:normal}, blank marks (-) indicate unavailable results.}

\vspace{-4pt}
\section{Experiments}
\label{sec:experiment}
%\todo[inline]{vxa: How about we first tell readers what is the objective of Experiments? (Presumably the goal is to to demonstrate that DRIFT is an efficient solution for our IoT scenario. We can do fine-grained classification at a fraction of pixels)} 
\noindent
Our goal is the IoT setup consisting of poorly resourced edge camera device communicating data to a cloud-server to perform fine-grained classification. To this end we show that DRIFT achieves state-of-art performance with significantly fewer high-acuity pixels relative to a fully trained DNN model that has access to full high-resolution images. %during training and test-time, while our proposed model must utilize a small number of pixels to perform the same task.

% We validated our formulation on a challenging task of fine-grained classification.
Experiments were conducted on five fine-grained classification datasets: CUB-200-2011~\cite{wah2011caltech}, Stanford Cars~\cite{KrauseStarkDengFei-Fei_3DRR2013}, Dogs~\cite{imagenet_cvpr09}, Aircrafts~\cite{maji13fine-grained}, and Food-101~\cite{bossard14}.
We chose these datasets since the distinctions among categories are subtle and highly local, which requires a foveation model to fixate on the most discriminative regions to classify  an image. The detailed statistics of these datasets are summarized in Table~\ref{table:data}. Only the image-level category labels were used for training, while extra annotations such as bounding boxes and parts were NOT used.

\begin{table}[h]
\centering
\renewcommand{\arraystretch}{1}
\setlength{\tabcolsep}{0.2cm}
\begin{tabular}{|l | c | c | c | c  | c | c | }
\whline
datasets  & CUB & Cars & Dogs & Air & Food \\
\hline
\# Category & 200 & 196 & 120 & 100 & 101\\
\# Train & 5,994 & 8,144 & 12,000 & 6,667 & 75,750 \\
\# Test & 5,794 & 8,041  & 8,580 & 3,333 & 25,250 \\
\whline 
\end{tabular}
%\vspace{-0.3cm}
\caption{The statistics of fine-grained datasets.}
\label{table:data}
\end{table}

% \begin{table*}[t]
% \begin{minipage}[b]{0.35\textwidth}
% \centering
% \renewcommand{\arraystretch}{1.2}
% \setlength{\tabcolsep}{0.1cm}
% \begin{tabular}{|l | c | c | c | }
% \whline
% datasets  & \# Class & \# Train & \# Test \\
% \hline
% CUB & 200 & 5994 & 5794\\
% Cars & 196 & 8144 & 8041\\
% Dogs & 120 & 12000 & 8580 \\
% Air & 100 & 6667  & 3333 \\
% Food & 101 & 75750 & 25250 \\
% \whline 
% \end{tabular}
% \caption{The statistics of fine-grained datasets. \label{table:data}}
% \end{minipage}\hfill
% \begin{minipage}[b]{0.65\textwidth}
% \centering
% \includegraphics[width=0.9\textwidth]{images/comparison.pdf}
% \captionof{figure}{Fixation comparison. \label{fig:comparison}}
% \end{minipage}
% \end{table*}

\begin{figure}[b]
    \centering
    \includegraphics[width=0.48\textwidth]{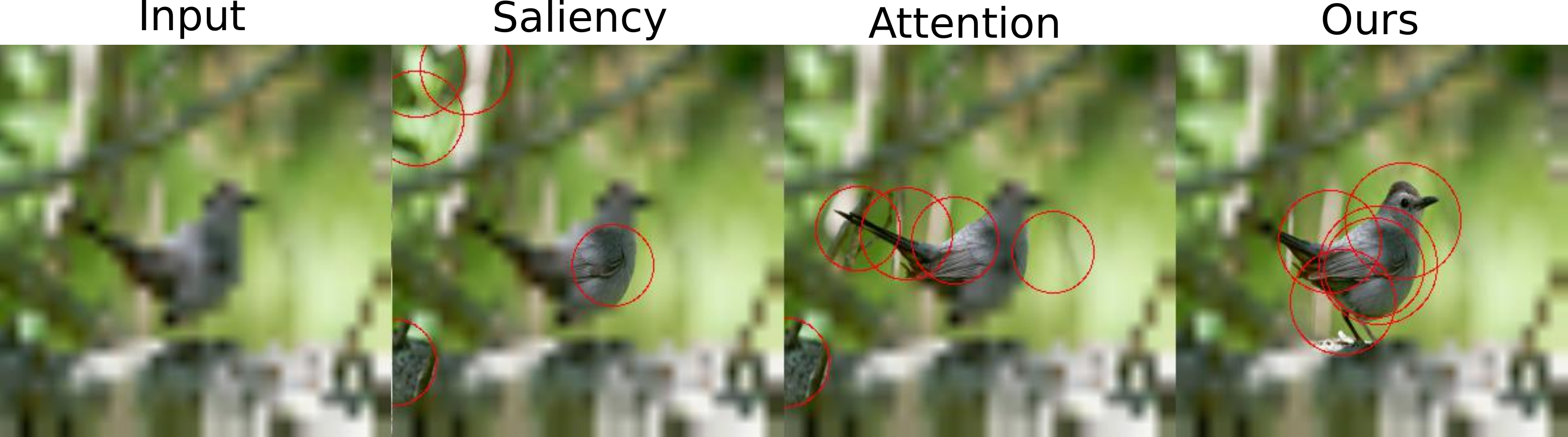}
    \caption{Comparison on mixed-acuity foveated images.}
    \label{fig:comparison}
\end{figure}

\begin{figure*}[t]
    \centering
    \includegraphics[width=0.95\textwidth]{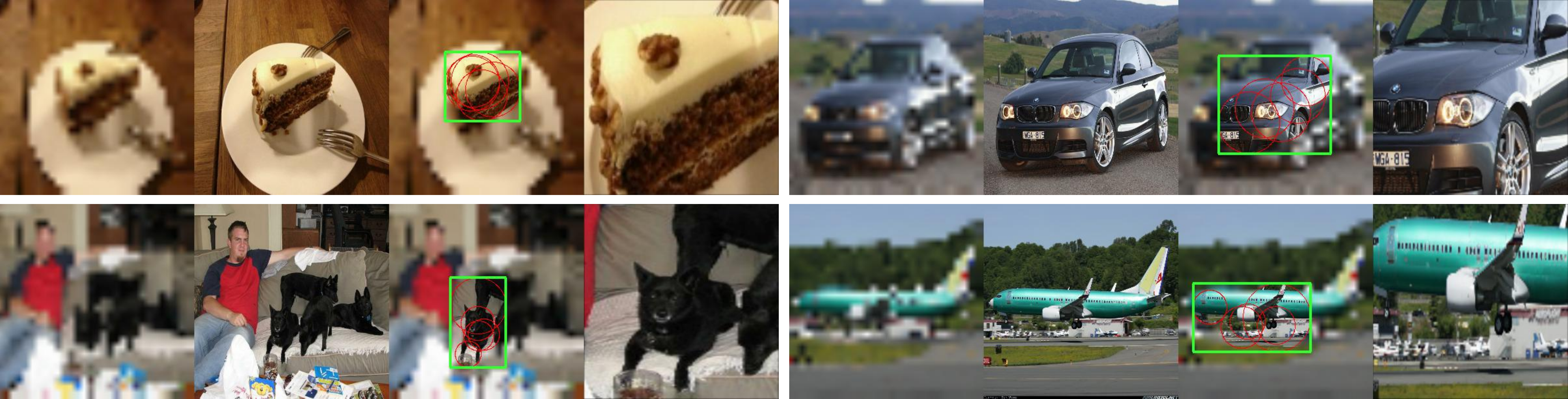}
    \caption{Qualitative result of the proposed DRIFT model. Each cell contains 4 images, from left to right: the (rescaled) low-acuity image $I_{low}$ (input), the high-acuity image $I_{high}$, DRIFT's foveated image $I_T$, and the zoomed high-acuity image by the tightest bounding box (shown in green) around the fixation points. On $I_T$, the fixation actions are also shown in red circles.}
    \vspace{-5pt}
    \label{fig:bboxes}
\end{figure*}
\vspace{-0.1in}
\subsection{Internet-of-Things Setting}
\label{sec:experiment_foveate}
\noindent \textbf{Setting} We first compare the proposed DRIFT under the IoT scenario as described in Sec.\ref{sec:foveation_for_classification}. Specifically, with the low-acuity images as initial inputs, we use DRIFT to generate fixation points under different foveation strategies,
acquire more high-acuity pixels from the edge device, 
and finally use the trained $f\circ g$ to classify the foveated images. 
For this IoT setting, two criteria are considered:
(1) the classification accuracy, and 
(2) the percentage of high-acuity pixels transmitted. 
A good foveation model should achieve high classification accuracy while requiring fewer high-acuity pixels.

\noindent \textbf{Baselines} Seven IoT foveation strategies are compared: 
{\bf (1)} {\it Random}: Fixate at random locations (uniform distribution); 
{\bf (2)} {\it Center}: Fixate at the image center; 
{\bf (3)} {\it Saliency}: Given an input image, we first obtain a class prediction $\hat{y}$, generate a class-response saliency map for $\hat{y}$ following~\cite{zhou2018weakly}, and then sample a fixation location based on the map values. The procedure is repeated for $T$ steps; 
{\bf (4)} {\it Attention}: We trained a multi-attention model, MA-CNN~\cite{zheng2017learning}, with $T$ parts. Given an input image $I_{low}$, it generates $T$ attention maps, from which $T$ fixation locations are sampled.
{\bf (5)} {\it BubbleNet}: BubbleNet~\cite{matzen2015bubblenet} initializes 128 fixation locations per image, iteratively optimizes each fixation and selects the best ones based on prediction entropy. We report its published results with the same Inception architecture;
{\bf (6)} {\it DRIFT}: Use the proposed model to generate sequential fixations;
{\bf (7)} {\it DRIFT$_E$}: In this strategy, if the prediction's entropy on our $I_T$ is higher than a threshold, we explore all high-acuity pixels in $I_{high}$ instead. The threshold is controlled so that only $25\%$ of the test images is used at full $I_{high}$.  
For (1-2), we control the fixation radius so that $15\%$ of high-acuity pixels are explored for easy comparisons. For (3-4), we randomly sample the fixation radius.\footnote{The proposed DRIFT model requires even less high-acuity pixels, so comparison is fair (see Table.\ref{table:foveat}).}

\noindent \textbf{Results} The results are shown in Table~\ref{table:foveat}. We also provide the direct classification results using $I_{low}$ and $I_{high}$ as context.
First, observe that DRIFT consistently outperforms the other five foveation strategies. While transmitting a similar number or fewer high-acuity pixels, DRIFT achieves much higher accuracy with our fixation approach. 
For example, on Aircrafts~\cite{maji13fine-grained} we achieve $86.7\%$ accuracy by only transmitting $14.4\%$ of the high-acuity pixels, only $0.5\%$ lower than the result with full high-acuity images ($87.2\%$). Moreover, with DRIFT$_E$ we are able to obtain an even higher accuracy ($88.0\%$).
This indicates that DRIFT's fixations indeed contain the most discriminative regions, which can be validated by Fig.\ref{fig:bboxes}. Taking a low-acuity image with limited information as input, it successfully fixates on objects of interest (e.g. the black dog), or the discriminative visual parts of an object (e.g. the headlight and grille of the BMW). 

Second, results in Table~\ref{table:foveat}  indicate that approaches like Saliency~\cite{zhou2018weakly} and Attention~\cite{zheng2017learning} fail to infer good fixations from low-acuity input (Fig.\ref{fig:comparison}), faring worse than Center fixations. This is because (a) they are not designed to operate on low/mixed-acuity inputs, and (b) they cannot accumulate prior fixations to inform future actions (DRIFT handles this via its proposed state representations and RL training guided by the dense rewards)
Third, observe that  on all datasets except for Food-101, Center fixation performs much better than Random fixation. This is because these datasets are artificially constructed by human with a center bias. For images in real-world deployments where the center prior no longer holds, we can expect a larger performance gap between DRIFT and Center fixations.

% \begin{table}[t]
% \centering
% \renewcommand{\arraystretch}{1}
% \setlength{\tabcolsep}{0.2cm}
% \begin{tabular}{|l | c  c  c  c  c|}
% \whline
% datasets  & CUB & Cars & Dogs & Air & Food \\
% \hline
% \hline
% Lin \etal.~\cite{lin2015bilinear} & 84.1 & 91.3 & - & 84.1 & - \\
% \rowcolor{Gray} Krause \etal.~\cite{krause2015fine} & 82.0 & 92.6 & 82.6 & - & - \\
% Fu \etal.~\cite{fu2017look} & 85.3 & 92.5 & 87.3 & 88.2 & - \\
% \rowcolor{Gray} Liu \etal.~\cite{liu2016fully} & 84.3 & 91.5 & \bf 88.9 & - & 86.3\\
% \hline
% Finetune (G)  &  81.6 & 91.2 & 81.8 & 87.2 & 85.0 \\
% \rowcolor{Gray} DRIFT (G)  &  83.7 & 92.2 & 82.9 & 90.7 & 86.6\\
% \hline 
% Finetune (R) & 83.2   & 92.2  & 85.7  & 89.8  & 85.8  \\
% \rowcolor{Gray} DRIFT (R)  & \bf 86.2  & \bf 93.6  &  87.3 & \bf 91.7  & \bf 88.6  \\
% \whline 
% \end{tabular}
% %\vspace{-0.3cm}
% \caption{Standard fine-grained classification results. (G: Inception-V3; R: ResNet-50)}
% \label{table:normal}
% \end{table}

\begin{figure*}[t]
    \centering
    \includegraphics[width=0.95\textwidth]{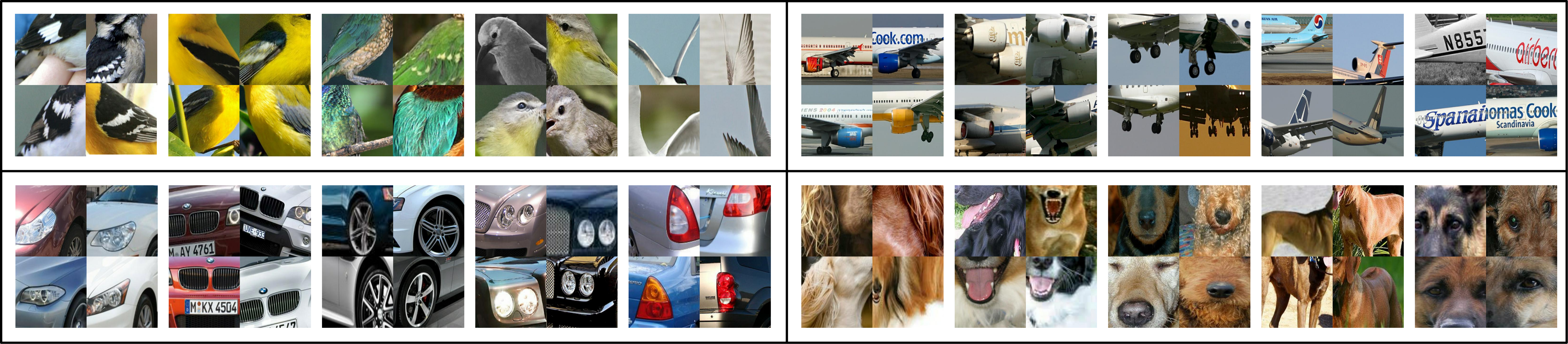}
    \caption{Visual patterns in fixation actions. Each cell contains four example fixation patches which belong to the same cluster. }
    \label{fig:clusters}
    \vspace{-15pt}
\end{figure*}

\subsection{Standard Setting}
\label{sec:class_normal}
\noindent \textbf{Setting and Baselines} 
Having validated DRIFT's ability in optimizing data transmission for IoT, 
we ask if DRIFT's fixation strategy is also beneficial to standard classification tasks.
Specifically, as shown in Fig.\ref{fig:bboxes}, we can fit a bounding box around DRIFT's fixations. The box is similar to a hard attention, with the only difference that it is generated via a sequential foveation procedure from a low-acuity input image.
In this setting, we simply treat DRIFT as a hard attention model, and verify whether it can boost classification results for any baseline classification model under the standard fine-grained classification setting. 

Specifically, we use DRIFT's hard attentions to zoom into the original images (Fig.\ref{fig:bboxes}). 
Given a baseline model, we simply fuse its predictions on the original image and the zoomed image by DRIFT's attention. We tested two baseline models: Inception-V3~\cite{szegedy2016rethinking} and ResNet-50~\cite{he2016deep}. 
For all the five datasets, they are pre-trained on ImageNet~\cite{imagenet_cvpr09}, and further trained for 30 epochs with a RMSProp optimizer and a batch size of 32. The learning rate is initialized as $0.01$ and decays $0.96$ every 4 epochs. The input sizes are $299$ and $448$ for Inception-V3 and ResNet-50, respectively.
We use DRIFT$_I$ and DRIFT$_R$ to represent the corresponding classification results using our hard attentions together with Inception-V3 and ResNet-50 baseline models.

\noindent \textbf{Results} Table~\ref{table:normal} shows our results. We observe a clear positive effect of DRIFT's attention selection on classification accuracy. In particular, on average DRIFT$_I$ is $1.9\%$ higher than Inception-V3 in absolute accuracy, while DRIFT$_R$ is $2.1\%$ higher than ResNet-50, and has already achieved better or comparable performance to existing state-of-the-arts. This again demonstrates DRIFT's ability to fixate on discriminative regions and filter out background clutter. 
% Importantly, compared to existing attention models for fine-grained classification, e.g.~\cite{fu2017look,liu2016fully,sun2018multi}, DRIFT's attentions are generated by exploring only very limited (on average $12.3\%$ in pixels) high-acuity data via its sequential fixation actions, making it more suitable for applications whose bottleneck is determined by transmission efficiency.
Note that to compare purely in accuracy between Table.\ref{table:foveat} and \ref{table:normal} is not meaningful since DRIFT and DRIFT$_E$ in Table.\ref{table:foveat} use far fewer pixels resulting in only around 10\% and 30\% pixel transmissions respectively while achieving accuracy very close to state-of-the-art as reported in Table.\ref{table:normal}.

\begin{table}[t]
\centering
\renewcommand{\arraystretch}{1} 
\setlength{\tabcolsep}{0.35cm}
\begin{tabular}{|l | c | c | c | c  | c  | }
\whline
datasets  & CUB & Cars & Dogs & Air  \\
\hline
\hline
Random & 8.0 & 34.8 & 25.6 & 11.9\\
Center & 36.2 & 89.1 & 61.2 & 32.9 \\
DRIFT & \bf 44.3 & \bf 91.5  & \bf 63.2 & \bf 50.9 \\
% ours (iou 0.5) & 22.8 & 8.0  & 11.1 & 8.9 \\
% \hline
% ours (iou 0.4) & 45.2 & 25.0 & 26.3 & 43.6  \\
% \hline
% ours (iou 0.3) & 72.3 & 58.7 & 52.8 & 83.5 \\
\whline 
\end{tabular}
%\vspace{-0.3cm}
\caption{Localization results in hit rate.}
\label{table:localizatoin}
\end{table}

\subsection{Discussion and Analysis}
\noindent \textbf{Where does DRIFT fixate?} 
First, 
%we aim to measure the correlation between DRIFT fixations to the locations of objects.
inspired by~\cite{zhang2018top} we use hit rate to evaluate the localization performance.
Specifically, taking the boxes generated by DRIFT as in Sec.~\ref{sec:class_normal}, we count a box as a hit when its intersection with the ground-truth box\footnote{CUB, Cars, Dogs and Aircrafts provide ground-truth bounding boxes.} is greater than $90\%$ of its own area, otherwise as a miss, and then measure $\frac{\# hits}{\# hits + \# misses}$.
The localization performance is shown in Table \ref{table:localizatoin}. We also show results of a randomly-generated box and a center-located box of $1/2$ image size. DRIFT's localization performance is consistently superior.
Evidently, DRIFT's fixations are strongly correlated to object locations, even though trained without location labels.

Second, we aim to discover and visualize common patterns in DRIFT's fixations to better understand its learned foveation policy. Specifically, we collect the local image patches specified by every fixation action, and perform a k-means clustering ($k=50$) over their visual features. The clusters with top popularity are shown in Fig.~\ref{fig:clusters}.
It is evident that DRIFT performs implicit part detection during fixation. This experiment also shows the potential applications of DRIFT in visual discovery.

\noindent \textbf{How much gain does `C3' provide?} While keeping all other settings fixed, we re-trained the foveation actor network $\pi_w$ with three different strategies: {DDPG}, { DDPG + Conditioned Critic}, and {DDPG + Coaching}. Table~\ref{table:ablative} shows the classification results on their foveated images (detailed test setting in Sec.\ref{sec:experiment_foveate}). The original actor-critic training scheme as in DDPG~\cite{lillicrap2017continuous} fails in our foveation problem, due to the reasons analyzed in Sec.\ref{sec:DDPGC3}, i.e. a global state-value function difficult to approximate by the critic, and less informative guides provided by a randomly initialized critic. By conditioning the critic on every training episode, on average the accuracy is improved by $8.0\%$ over the three datasets. Moreover, by coaching the actor using the policy sampled from a heuristic oracle that reduces exploration efforts, on average a $19.0\%$ performance gain is obtained. Finally, the full DRIFT model, trained with the proposed DDPGC3 algorithm, brings a $25.2\%$ average improvement in absolute accuracy; clearly DDPGC3 trains a better foveation policy.

\begin{table}[t]
\centering
\renewcommand{\arraystretch}{1}
\setlength{\tabcolsep}{0.15cm}
\begin{tabular}{ | l |c | c | c | c | }
\whline
acc (\%) & DDPG & + con. critic  & + coaching & DRIFT \\
\hline
\hline
CUB  & 51.0 &  57.1 &  67.0 & \bf 74.4 \\
Cars & 48.3 &  61.4 &  76.6 & \bf 82.8 \\
Dogs & 53.8 &  58.5 &  66.6 & \bf 71.6 \\
\whline 
\end{tabular}
%\vspace{-0.3cm}
\caption{Ablative analysis with results on foveated images.}
\label{table:ablative}
\end{table}

\vspace{-10pt}
\section{Conclusion}
\vspace{-5pt}
We considered IoT scenarios where the cost of transmitting high-acuity images
from an edge device to the cloud exceeds the transmission/power budget.
Our solution is DRIFT, a novel deep-RL approach to generate sequential fixations with a foveated field-of-view. DRIFT avoids discretizing  the state-action space, which would be prohibitively expensive, and instead solves a continuous-control problem. As part of our solution we introduce a novel use of a {\em conditioned critic} and a {\em coaching} strategy; we also provide an example of shaping the reward function to accelerate convergence. Experiments show high accuracy and data-efficiency of our approach on challenging classification tasks. Finally, although we demonstrated the proposed model's effectiveness in classification, DRIFT is a general active image exploration solution which can be applied to other domains.

{\small
\bibliographystyle{ieee_fullname}
\bibliography{main.bbl}
}

\end{document}